\documentclass{article}

\usepackage[final,nonatbib]{neurips_2019}

\usepackage[utf8]{inputenc} 
\usepackage[T1]{fontenc}    
\usepackage{hyperref}       
\usepackage{url}            
\usepackage{booktabs}       
\usepackage{amsfonts}       
\usepackage{nicefrac}       
\usepackage{microtype}      

\usepackage{amsmath,amsthm,amssymb,euscript,stmaryrd,graphicx,enumitem,yhmath,mathrsfs}
\usepackage{algorithmic,algorithm,mathtools}
\usepackage{bm}

\usepackage{xcolor}
\definecolor{codegreen}{rgb}{0,0.6,0}
\definecolor{codegray}{rgb}{0.5,0.5,0.5}
\definecolor{codepurple}{rgb}{0.58,0,0.82}
\definecolor{backcolour}{rgb}{0.95,0.95,0.92}

\title{Biologically inspired architectures for sample-efficient deep reinforcement learning}

\author{%
 Pierre H. Richemond\\
 Imperial College London\\
 \texttt{phr17@imperial.ac.uk} \\
 \And
 Arinbjörn Kolbeinsson\\
 Imperial College London\\
 \texttt{ak711@imperial.ac.uk} \\
 \And
Yike Guo\\
 Imperial College London\\
 \texttt{y.guo@imperial.ac.uk} \\
}

\begin{document}

\maketitle

\begin{abstract}
Deep reinforcement learning requires a heavy price in terms of sample efficiency and overparameterization in the neural networks used for function approximation. In this work, we use \emph{tensor factorization} in order to learn more compact representation for reinforcement learning policies. We show empirically that in the low-data regime, it is possible to learn online policies with 2 to 10 times less total coefficients, with little to no loss of performance. We also leverage progress in second order optimization, and use the theory of \emph{wavelet scattering} to further reduce the number of learned coefficients, by foregoing learning the topmost convolutional layer filters altogether. We evaluate our results on the Atari suite against recent baseline algorithms that represent the state-of-the-art in data efficiency, and get comparable results with an order of magnitude gain in weight parsimony.
\end{abstract}

\section{Introduction \& Related Work}

The successes of deep reinforcement learning (thereafter 'RL') come at a heavy computational price. It is well known that achieving human-level performance in domains such as Atari \cite{suttonbarto, dqn, rainbow} requires hundreds of millions of frames of environment interaction. As such, the problem of sample efficiency in reinforcement learning is of critical importance. Several tracks of concurrent research are being investigated, and have reduced by orders of magnitude the number of environment interactions required for good performance beyond the previous benchmark of biologically-inspired episodic control methods \cite{mfec,nec} to a couple hours of human gameplay time \cite{dataefficientrainbow, simple}.

However, while the data-efficiency of RL methods has seen recent drastic performance, their function approximators still use millions of learned weights, potentially still leaving them heavily overparameterized. Independently motivated by biological facts like the behavioural readiness of newborn animals, several authors \cite{weightagnostic,sixneurons, red} have recently looked at doing away with learning so many weights for RL tasks. Smaller networks not only train faster, but may yet offer another avenue for gains in the form of better generalization \cite{understanding}. Very recent work from \cite{weightagnostic} studies the effect of inductive bias of neural architectures in reinforcement learning ; they forego training altogether, but transfer networks that only obtain 'better than chance performance on MNIST'. In similar fashion, \cite{red} investigate the effect of random projections in the restricted setting of imitation learning. Finally, \cite{sixneurons} manage human-level performance on the Atari suite using a separate dictionary learning procedure for their features, bypassing the usual end-to-end learning paradigm. The perspective of neural architecture search applied to RL appears difficult, if not computationally inextricable.

Concurrently, the study of biologically-inspired models of learning has exhibited two mathematical characterizations that might be critical in explaining how biological learning takes place so efficiently. First, the low-rank properties of learned perceptual manifolds \cite{sompolinskyclassification, sompolinskyreadout} are giving rise to a rich theory borrowing from statistical physics. Second, another well known line of work has identified Gabor filters (and more generally wavelet filter-like structures) in the actual visual cortex of animals \cite{jones1987}, and linked those to sparsity-promoting methods and dictionary learning \cite{olshausenemergence, olshausensparse, hyvrinen}. But these breakthroughs have not, so far, been reflected as inductive priors in the shape of modifications in deep RL neural networks architectures, which remain fairly fixed on the Atari domain.

Therefore the following questions remain: how parsimonious do function approximators in reinforcement learning need to be, in order to maintain good performance? And can we be at once \emph{sample-efficient} and \emph{weight-efficient} ? In this work, we turn to the mathematical theories of tensor factorization \cite{Cichocki}, second-order optimization \cite{amarinatural, KFAC} and wavelet scattering \cite{mallatscattering} to answer this question positively and empirically, in a model-free setting. To the best of our knowledge, this is the first time those fields have been combined together in this context, and that tensor factorization is applied to deep RL.

\section{Background}

\subsection{Deep Reinforcement Learning}
We consider the standard Markov Decision Process framework as in \cite{suttonbarto}).  This setting is characterised by a tuple $\langle S,A,T,R,\gamma \rangle$, where $S$ is a set of states, $A$ a set of actions, $R$ a reward function  that is the immediate, intrinsic desirability of a certain state, $T$ a transition dynamics and $\gamma$ $\in [0,1]$ a discount factor. The purpose of the RL problem is to to find a policy $\pi$, which represents a mapping from states to a probability distribution over actions, that is optimal, i.e., that maximizes the expected cumulative discounted return $\sum_{k=0}^{\infty} \gamma ^{k}R_{t+k+1}$ at each state $s_t \in S$.  In Q-learning, the policy is given implicitly by acting greedily or $\epsilon$-greedily with respect to learned \emph{action-value functions} $q^{\pi}(s,a)$, that are learned following the Bellman equation. In \emph{deep Q-learning}, $q_{\theta}$ becomes parameterized by the weights $\theta$ of a neural network and one minimizes the expected Bellman loss :
$$\mathbb{E} \left(R_{t+1}+\gamma_{t+1} \max _{a^{\prime}} q_{\theta}\left(S_{t+1}, a^{\prime}\right)-q_{\theta}\left(S_{t}, A_{t}\right)\right)^{2}$$
In practice, this is implemented stochastically via uniform sampling of transitions in an experience replay buffer, as is done in the seminal paper \cite{dqn}. Several algorithmic refinements to that approach exist. First, Double Q-learning \cite{doubledqn} proposes to decouple learning between two networks in order to alleviate the Q-value overestimation problem. Second, \emph{dueling} Q-networks \cite{duelingdqn} explicitly decompose the learning of an action-value function $q_{\theta}(s,a)$ as the sum of an action-independent state-value, much like what is traditionally done in policy gradient methods \cite{suttonbarto}, implemented via a two-headed neural network architecture. Finally, \emph{prioritized} RL \cite{prioritized} proposes to replace the uniform sampling of transitions in the experience replay buffer with importance sampling, by prioritizing those transitions that present the most Bellman error (those transitions that are deemed the most 'surprising' by the agent). \cite{noisy} uses extra weights to learn the variance of the exploration noise in a granular fashion, while \cite{c51} proposes to learn a full \emph{distribution} of action-values for each action and state. Combined, those methods form the basis of the Rainbow algorithm in \cite{rainbow}.

\subsection{Tensor factorization}

Here we introduce notations and concepts from the tensor factorization literature. An intuition is that the two main decompositions below, \emph{CP} and \emph{Tucker} decompositions, can be understood as multilinear algebra analogues of SVD or eigendecomposition. \\

\textbf{CP decomposition.} A tensor $\mathcal{X} \in \mathbb{R}^{I_{1} \times I_{2} \times \cdots \times I_{N}}$, can be decomposed into a sum of $R$ rank-1 tensors, known as the Canonical-Polyadic decomposition, where $R$ is as the rank of the decomposition. The objective is to find the vectors $\mathbf{u}_{k}^{(1)}, \mathbf{u}_{k}^{(2)}, \cdots, \mathbf{u}_{k}^{(N)}$, for $k=[1 \ldots R]$, as well as a vector of weights $\boldsymbol{\lambda} \in \mathbb{R}^{R}$ such that:
$$\mathcal{X}=\sum_{k=1}^{R} \underbrace{\lambda_{k} \mathbf{u}_{k}^{(1)} \circ \mathbf{u}_{k}^{(2)} \circ \cdots \circ \mathbf{u}_{k}^{(N)}}_{\text { rank-1 } \text { components }}$$

\textbf{Tucker decomposition.} A tensor $\mathcal{X} \in \mathbb{R}^{I_{1} \times I_{2} \times \cdots \times I_{N}}$, can be decomposed into a low rank approximation consisting of a core $\mathcal{G} \in \mathbb{R}^{R_{1} \times R_{2} \times \cdots \times R_{N}}$ and a set of projection factors $\left(\mathbf{U}^{(0)}, \cdots, \mathbf{U}^{(N-1)}\right)$, with  $\mathbf{U}^{(k)} \in \mathbb{R}^{R_{k}, \hat{I}_{k}}, k \in(0, \cdots, N-1)$ that, when projected along the corresponding dimension of the core, reconstruct the full tensor $\mathcal{X}$. The tensor in its decomposed form can then be written:
$$\begin{aligned} \mathcal{X} &=\mathcal{G} \times_{1} \mathbf{U}^{(1)} \times_{2} \mathbf{U}^{(2)} \times \cdots \times_{N} \mathbf{U}^{(N)} = \left[\mathcal{G} ; \mathbf{U}^{(1)}, \cdots, \mathbf{U}^{(N)}\right]\end{aligned}$$

\textbf{Tensor regression layer.} For two tensors $\mathcal{X} \quad \in \quad \mathbb{R}^{{K}_{1} \times \cdots \times K_{x} \times I_{1} \times \cdots \times I_{N}}$ and $\mathcal{Y} \quad \in \quad \mathbb{R}^{I_{1} \times \cdots \times I_{N} \times L_{1} \times \cdots \times L_{y}}$, we denote by $\langle\mathcal{X}, \mathcal{Y}\rangle_{N} \quad \in \quad \mathbb{R}^{K_{1} \times \cdots \times K_{x} \times L_{1} \times \cdots \times L_{y}}$ the contraction of $\mathcal{X}$ by $\mathcal{Y}$ along their $N$ last modes; their generalized inner product is
$$\langle\mathcal{X}, \mathcal{Y}\rangle_{N} = \sum_{i_{1}=1}^{I_{1}} \sum_{i_{2}=1}^{I_{2}} \cdots \sum_{i_{n}=1}^{I_{N}} \mathcal{X}_{\ldots, i_{1}, i_{2}, \ldots, i_{n}} \mathcal{Y}_{i_{1}, i_{2}, \ldots, i_{n}, \ldots}$$
This enables us to define a \emph{tensor regression layer} \cite{tensorregressionnetworks} that is differentiable and learnable end-to-end by gradient descent. Let us denote by $\mathcal{X} \in \mathbb{R}^{I_{1} \times I_{2} \times \cdots \times I_{N}}$ the input activation tensor for a sample and $\mathbf{y} \in \mathbb{R}^{I_{N}}$ the label vector. A tensor regression layer estimates the regression weight tensor $\mathcal{W} \in \mathbb{R}^{I_{1} \times I_{2} \times \cdots \times I_{N}}$ under a low-rank decomposition. In the case of a Tucker decomposition (as per our experiments) with ranks $\left(R_{1}, \cdots, R_{N}\right)$, we have :
$$\begin{array}{c}{\mathbf{y}=\langle\mathcal{X}, \mathcal{W}\rangle_{N}+\mathbf{b}} \qquad {\text { with } \mathcal{W}=\mathcal{G} \times_{1} \mathbf{U}^{(1)} \times_{2} \mathbf{U}^{(2)} \cdots \times_{N} \mathbf{U}^{(N)}}\end{array}$$ 
as  $\mathcal{G} \in \mathbb{R}^{R_{1} \times \cdots \times R_{N}}$, $\mathbf{U}^{(k)} \in \mathbb{R}^{I_{k} \times R_{k}}$ for each $k$ in $[1 \ldots N]$ and $\mathbf{U}^{(N)} \in \mathbb{R}^{1 \times R_{N}}$.

\cite{tensorcontractionlayers, tensorregressionnetworks, trllr} learn parsimonious deep learning fully-connected layers thanks to low-rank constraints.

\subsection{Wavelet scattering} The \emph{wavelet scattering transform} was originally introduced by \cite{mallatscattering} and \cite{brunamallat} as a non-linear extension to the classical wavelet filter bank decomposition \cite{Mallat1998AWT}. Its principle is as follows. Denoting by $x \circledast y[n]$ the 2-dimensional, circular convolution of two signals $x[n]$ and $y[n]$, let us assume that we have pre-defined two wavelet filter banks available $\left\{\psi_{\lambda_{1}}^{(1)}[n]\right\}_{\lambda_{1} \in \Lambda_{1}}$ $\left\{\psi_{\lambda_{2}}^{(2)}[n]\right\}_{\lambda_{2} \in \Lambda_{2}}$ , with $\lambda_1$ and $\lambda_2$ two frequency indices. These wavelet filters correspond to high frequencies, so we also give ourselves the data of a lowpass filter $\phi_{J}[n]$. Finally, and by opposition to traditional linear wavelet transforms, we also assume a given nonlinearity $\rho(t)$. Then the scattering transform is given by coefficients of order 0,1, and 2, respectively :
$$S_{0} x[n] = x \circledast \phi_{J}[n] $$
$$S_{1} x\left[n, \lambda_{1}\right]=\rho\left(x \circledast \psi_{\lambda_{1}}^{(1)}\right) \circledast \phi_{J}[n] \quad \lambda_{1} \in \Lambda_{1}$$
$$S_{2} x\left[n, \lambda_{1}, \lambda_{2}\right]=\rho\left(\rho\left(x \circledast \psi_{\lambda_{1}}^{(1)}\right) \circledast \psi_{\lambda_{2}}^{(2)}\right) \circledast \phi_{J}[n] \quad \lambda_{1} \in \Lambda_{1}, \lambda_{2} \in \Lambda_{2}\left(\lambda_{1}\right)$$

This can effectively be understood and implemented as a two-layer convolutional neural network whose weights are not learned but rather frozen and given by the coefficients of wavelets $\psi$ and $\phi$ (with Gabor filters as a special case \cite{Mallat1998AWT}). The difference with traditional filter banks comes from the iterated modulus/nonlinear activation function applied at each stage, much like in traditional deep learning convolutional neural networks. In practice, the potential of scattering transforms to accelerate deep learning by providing ready-made convolutional layers weights has been investigated in \cite{oyallonfirstscattering, oyallonappliedscattering, ganscattering}.

\subsection{Second order optimization with K-FAC}

While stochastic gradient descent is usually performed purely from gradient observations derived from auto-differentiation, faster, second order optimization methods first multiply the weights' $\theta$ gradient vector  $\nabla_{\theta}$ by a preconditioning matrix, yielding the weight update $\theta \leftarrow \theta - \eta G^{-1} \nabla_{\theta}$. In the case of second order methods, the matrix $G^{-1}$ is chosen to act as a tractable iterative approximation to the inverse Hessian or Empirical Fisher Information Matrix \cite{amarinatural} of the neural network model in question. Kronecker-factored approximate curvature or K-FAC \cite{KFAC} enforces a Kronecker decomposition of the type $G = A \otimes B$, with $A$ and $B$ being smaller, architecture-dependent matrices. Unlike the above methods, K-FAC \emph{has} been applied as a plug-in in the RL literature and been shown to promote convergence \cite{ACKTR}.

\section{Methods \& Experimental Results}
We do take as a baseline method the data-efficient Rainbow of \cite{dataefficientrainbow}. However, we change the architecture of the neural network function approximators used, in accordance with the principles described above, combining them to reflect inductive biases promoting fewer learnable parameters:
\begin{itemize}
    \item We replace the fully-connected, linear layers used in the Rainbow \cite{rainbow} and data-efficient Rainbow \cite{dataefficientrainbow} by tensor regression layers \cite{tensorregressionnetworks} in order to learn \emph{low-rank} policies (ranks in appendix).
    \item We use either the K-FAC \cite{KFAC} second order stochastic optimizer, or ADAM \cite{adam}. 
    \item We combine the two methods with various rank and therefore weight compression ratios and evaluate those on the same subset of Atari games as \cite{ dataefficientrainbow, simple}.
    \item When possible, we replace the first convolutional layer in the approximating neural network with a \emph{scattering} layer for further gains in terms of learnable weights.
\end{itemize}

For all our Atari experiments, we used OpenAI Gym \cite{openaigym}, and a combination of PyTorch \cite{pytorch}, TensorLy \cite{tensorly} and Kymatio \cite{kymatio} for auto-differentiation.
We evaluated our agents in the low-data regime of 100,000 steps, on half the games, with 3 different random seeds for reproducibility \cite{drlthatmatters}. Our specific hyperparameters are described in appendix. We report our results in tables \ref{table1} and \ref{table2}.
\begin{table}[ht!]
\centering
\begin{tabular}{lrrrrrr}
\toprule
            Game &    \textcolor{blue}{SimPLe} &  \textcolor{blue}{Rainbow} & Denoised & TRL 2.5x & TRL 5x & TRL 10x \\
\midrule
           alien &    405 &    \textbf{740} & 684 & 688 & 454 & 566 \\
          amidar &    88 &    \textbf{189} & 154 & 118 & 86 & 84\\
         assault &     369 &    431 & 321 & \textbf{543} & 521 & 513\\
         asterix &   \textbf{1090} &    471 & 500 & 459 & 554 & 363\\
      bank\_heist &        8 &     51 & 77 & 59 & \textbf{134} & 42\\
     battle\_zone &   5184 &  10125 & 9378 & \textbf{14466} & 13466 & 5744\\
          boxing &       \textbf{9} &      0.2 & 1 & -2 & -2 & -5\\
        breakout &       \textbf{13} &      2 & 3 & 2 & 2 & 4\\
 chopper\_command &    1247 &    862 & \textbf{1293} & 1255 & 1243 & 1106\\
   crazy\_climber &  \textbf{39828} &  16185 & 9977 & 3928 & 4225 & 2340\\
    demon\_attack &     170 &    \textbf{508} & 450 & 362 & 263 & 175\\
         freeway &      20 &     \textbf{28} & 28 & 26 & 25 & 24\\
       frostbite &   255 &    867 & \textbf{1101} & 659 & 912 & 231\\
          gopher &    \textbf{771} &    349 & 391 & 278 & 255 & 396\\
            hero &  1295 &   \textbf{6857} & 3013 & 5351 & 3732 & 3321\\
       jamesbond &     125 &    \textbf{302} & 295 & 215 & 213 & 218\\
        kangaroo &     323 &    \textbf{779} & 1002 & 804 & 715 & 400\\
           krull &   \textbf{4540} &   2852 & 2656 & 2333 & 2275 & 2308\\
  kung\_fu\_master &  \textbf{17257} &  14346 & 4037 & 9392 & 4764 & 4031\\
       ms\_pacman &    763 &   \textbf{1204} & 1053 & 818 & 838 & 517\\
            pong &       \textbf{5} &    -19 & -20 & -20 & -19 & -21\\
     private\_eye &    58 &   98 & 100 & 51 & 100 & \textbf{1128}\\
           qbert &    560 &   \textbf{1153} & 672 & 697 & 581 & 733\\
     road\_runner &    5169.4 &   \textbf{9600} & 5426 & 6965 & 3914 & 1319\\
        seaquest &   371 &    354 & \textbf{387} & 345 & 350 & 287\\
       up\_n\_down &   2153 &  2877 & \textbf{5123} & 2197 & 2302 & 2179\\
\bottomrule
\textbf{Average (vs. Rainbow) } &  & \textbf{100\%} & \textbf{118\%} & \textbf{96\%} & \textbf{90\%} & \textbf{71\%} \\
\vspace{-2pt}
\end{tabular}
\caption{Mean episode returns as reported in \textcolor{blue}{SimPLe} \cite{simple} and \textcolor{blue}{data-efficient Rainbow} \cite{dataefficientrainbow}, versus our agents, on 26 Atari games. \emph{'Denoised'} is the NoisyNet ablation of Rainbow; \emph{'TRL'} shows the performance of the data-efficient Rainbow with tensor regression layers substituted for linear ones.}
\label{table1}
\end{table}

Table \ref{table1} shows proof of concept of the online learning of low-rank policies, with a loss of final performance varying in proportion to the compression in the low-rank linear layers, very much like in the deep learning literature \cite{tensorcontractionlayers, tensorregressionnetworks}. The number of coefficients in the original data-efficient Rainbow is of the order of magnitude of 1M and varies depending on the environment and its action-space size. The corresponding tensor regression layer ranks are in appendix, and chosen to target 400k, 200k and 100k coefficients respectively. While individual game results tend to decrease monotonously with increasing compression, we observe that they are noisy as per the nature of exploration in RL, and average scores reported correspond with the intuition that performance seems to decrease fast after a certain overparameterization threshold is crossed. To take this noisy character into account, we take care to be conservative and report the average of the final three episodes of the learned policy after 80000, 90000 and 100000 steps, respectively. Also, so as to not muddy the discussion and provide fair baselines, we do report on the NoisyNet \cite{noisy} ablation of Rainbow ('Denoised' columns), as the NoisyLinear layer doubles up the number of coefficients required and actually performs worse in our experiments. Interestingly, the approximation error in tensor factorization seems to play a role akin to promoting exploration noise.

We then proceed to assess the impact of second-order optimization to our architecture by substituting ADAM optimization for K-FAC, and introducing scattering, in table \ref{table2} (only a handful results being available with scattering, due to computational limitations). In spite of our conservative reporting, the efficiency boost from using a second order scheme more than makes up for low-rank approximation error with five times less coefficients than \cite{dataefficientrainbow}, suggesting that learning with a full order of magnitude less coefficients is well within reach of our techniques.

\begin{table}[ht!]
\centering
\begin{tabular}{lrrrr}
\toprule
            Game &      KFAC+Denoised & KFAC+TRL5x & KFAC+TRL10x & Scattering\\
\midrule
           alien &       \textbf{996} & 734 & 643 & 441  \\
          amidar &          \textbf{163} &  101 & 98 & 84 \\
         assault &         \textbf{501} & 491 & 496 & 434  \\
         asterix &          537 & \textbf{549} & 526 & 502 \\
      bank\_heist &              \textbf{100} &  73 & 57 & 29  \\
     battle\_zone &           8622  & \textbf{15178} & 6156 & 4311 \\
          boxing &                \textbf{0} & -4 & -1 & -9  \\
        breakout &              3 & \textbf{3} & 2 & 2  \\
 chopper\_command &            692 & 611 & \textbf{1302} & 441 \\
   crazy\_climber &         \textbf{14242} & 12377 & 3546 & 740 \\
    demon\_attack &             582 & 434 & 318 & \textbf{692} \\
         freeway &                26 & 26 & 24 & 19\\
       frostbite &         \textbf{1760} & 718 & 1483 & 654 \\
          gopher &         \textbf{363} & 341 & 265 & 172\\
            hero &     4188 & \textbf{6284} & 4206  & 4127 \\
       jamesbond &       263 & \textbf{327} & 217 & 48\\
        kangaroo &        \textbf{2085} & 613 & 588 & 391  \\
           krull &       2855 & \textbf{3441} & 3392 & 772 \\
  kung\_fu\_master &      8481 & \textbf{10738} & 7357 & 233   \\
       ms\_pacman &        \textbf{1137}  & 920 & 867 & 613   \\
            pong &        -19.3 & \textbf{-19} & -19 & -20    \\
     private\_eye &      56  & \textbf{100} & 100 & 0   \\
           qbert &        \textbf{731}   &  520 & 538 & 475  \\
     road\_runner &      4516 &  \textbf{8493} & 7224 & 1278  \\
        seaquest &       349 & 317 & \textbf{520} & 213 \\
       up\_n\_down &      \textbf{2557}  &  2291 & 2108 & 993  \\
\bottomrule
\textbf{Average (vs. Rainbow)} & \textbf{114\%} & \textbf{109\%} & \textbf{98\%} & \textbf{56\%} \\

\vspace{-8pt}
\end{tabular}
\caption{Mean episode returns of our low-rank agents with second-order optimization and scattering.}
\label{table2}
\end{table}
\vspace{-18pt}
\section{Conclusion}
\vspace{-6pt}
We have demonstrated that in the low-data regime, it is possible to leverage biologically plausible characterizations of experience data (namely low-rank properties and wavelet scattering separability) to exhibit architectures that learn policies with many times less weights than current baselines, \emph{in an online fashion}. We do hope that this will lead to even further progress towards sample efficiency and speedy exploration methods. Further work will first focus on thorough evaluation and research of scattering architectures in order to achieve further gains, and second investigate additional, orthogonal biologically-friendly research directions such as promoting sparsity.  


\medskip
\small


\newpage
\bibliography{neurips_2019}
\bibliographystyle{ieeetr}

\newpage
\section*{Appendix}

\subsection*{Hyperparameters and Reproducibility}

Our codebase is available on request. Hyperparameters are as follows. First, our specific architecture-modified hyperparameters:

\begin{table}[ht]
\small
\centering
\begin{tabular}{lrr}
\toprule
Specific architecture hyperparameters & \multicolumn{2}{r}{Value} \\
\midrule
Scattering maximum log-scale $J$ && 3 \\
Scattering volume width $M$ && 1\\
Scattering tensor input shape && (1,4,84,84) \\
Scattering tensor output shape && (1,16,11,11)\\
Scattering type && Harmonic 3D, see \cite{kymatio, harmonicscattering} \\
\midrule
Hidden linear layer rank constraint, 2.5x compression && 128 \\
Final linear layer rank constraint, 2.5x compression && 48 \\
Hidden linear layer rank constraint, 5x compression && 32 \\
Final linear layer rank constraint, 5x compression && 48 \\
Hidden linear layer rank constraint, 10x compression && 16 \\
Final linear layer rank constraint, 10x compression && 10 \\
\midrule
KFAC Tikhonov regularization parameter && 0.1 \\
KFAC Update frequency for inverses && 100 \\
\bottomrule
\end{tabular}
\vspace{2pt}
\caption{Our additional, architecture-specific hyperparameters.}
\label{table4}
\end{table}

Furthermore, we mirror the Data-Efficient Rainbow \cite{dataefficientrainbow} baseline:

\begin{table}[ht]
\small
\centering
\begin{tabular}{lrr}
\toprule
Data-efficient Rainbow hyperparameters & \multicolumn{2}{r}{Value} \\
\midrule
Grey-scaling && True \\
Observation down-sampling && (84, 84) \\
Frames stacked && 4 \\
Action repetitions && 4 \\
Reward clipping && [-1, 1] \\
Terminal on loss of life && True \\
Max frames per episode && 108K \\
Update & \multicolumn{2}{r}{Distributional Double Q} \\
Target network update period${}^{*}$ & \multicolumn{2}{r}{every 2000 updates} \\
Support of Q-distribution && 51 bins \\
Discount factor && 0.99 \\
Minibatch size && 32 \\
Optimizer && Adam \\
Optimizer: first moment decay && 0.9 \\
Optimizer: second moment decay && 0.999 \\
Optimizer: $\epsilon$ && $0.00015$ \\
Max gradient norm && 10 \\
Priority exponent && 0.5 \\
Priority correction${}^{**}$ && 0.4 $\rightarrow$ 1\\
Hardware && NVidia 1080Ti GPU \\
Noisy nets parameter && 0.1 \\

Training frames &&  400,000 \\
Min replay size for sampling &&  1600 \\
Memory size &&  unbounded \\
Replay period every && 1 steps \\
Multi-step return length && 20 \\
Q network: channels && 32, 64 \\
Q network: filter size && 5 x 5, 5 x 5 \\
Q network: stride  && 5, 5 \\
Q network: hidden units && 256 \\
Optimizer: learning rate && 0.0001 \\
\bottomrule
\end{tabular}
\vspace{2pt}
\caption{Data-efficient Rainbow agent hyperparameters, as per \cite{dataefficientrainbow}.}
\label{table3}
\end{table}

\subsection*{Standard deviations for score runs}

\begin{table}[ht!]
\centering
\begin{tabular}{lrrrr}
\toprule
            Game &          Denoised & TRL 2.5x &  TRL 10x\\
\midrule
           alien &         684  $\pm$   7 & 688 $\pm$ 123 & 566$\pm$ 38  \\
          amidar &            154  $\pm$  21 & 118 $\pm$ 12 & 84$\pm$ 15 \\
         assault &        321  $\pm$ 224 & 543 $\pm$ 94 & 513$\pm$ 64  \\
         asterix &           500  $\pm$ 124 & 459 $\pm$ 91 & 363 $\pm$ 66\\
      bank\_heist &             77  $\pm$ 23  & 59 $\pm$ 22 & 42 $\pm$ 2  \\
     battle\_zone &           9378  $\pm$ 2042 &14466 $\pm$ 2845 & 5744 $\pm$ 575 \\
          boxing &                 1    $\pm$ 2 & -2 $\pm$ 1  & -5 $\pm$ 1 \\
        breakout &              3    $\pm$ 1.5 & 2 $\pm$ 1 & 4 $\pm$ 0.3    \\
 chopper\_command &          1293   $\pm$ 445 & 1255 $\pm$ 215 & 1106 $\pm$ 124  \\
   crazy\_climber &       9977 $\pm$ 3744 & 3928 $\pm$ 221 & 2340 $\pm$ 595 \\
    demon\_attack &            450    $\pm$ 49  &  362 $\pm$ 147 & 175 $\pm$ 7 \\
         freeway &                28    $\pm$ 0.6 & 26 $\pm$ 0 & 24 $\pm$ 0.5 \\
       frostbite &            1101   $\pm$ 355 & 659 $\pm$ 523 & 231 $\pm$ 1 \\
          gopher &          391    $\pm$ 46 & 278 $\pm$ 39 &  396 $\pm$ 24 \\
            hero &        3013    $\pm$ 90 & 5351 $\pm$ 1948 & 3321 $\pm$ 598 \\
       jamesbond &           295    $\pm$ 57  &215 $\pm$ 42 & 218 $\pm$ 22  \\
        kangaroo &         1002  $\pm$ 587 & 804 $\pm$ 289 & 400 $\pm$ 278   \\
           krull &       2656  $\pm$ 180 & 2333 $\pm$ 309 & 2308 $\pm$ 268   \\
  kung\_fu\_master &    4037  $\pm$ 2962 & 9392 $\pm$ 6289 & 4031 $\pm$ 3068    \\
       ms\_pacman &       1053   $\pm$ 193 & 818 $\pm$ 94 & 517 $\pm$ 38    \\
            pong &        -20   $\pm$ 0.4 & -20 $\pm$ 0 & -21 $\pm$ 0.1       \\
     private\_eye &      100   $\pm$ 0 & 51 $\pm$ 59 & 1128 $\pm$ 1067    \\
           qbert &      672   $\pm$ 144 & 697 $\pm$ 78 & 733 $\pm$ 291    \ \\
     road\_runner &     5426  $\pm$ 2830 & 6965 $\pm$ 6569  & 1319$\pm$ 216  \\
        seaquest &       387    $\pm$ 24 & 345 $\pm$ 40 & 287 $\pm$ 87 \\
       up\_n\_down &     5123  $\pm$ 3146 & 2197 $\pm$ 231 & 2179 $\pm$ 178   \\
\bottomrule
\vspace{2pt}
\end{tabular}
\caption{Standard deviations across seeds for runs presented Table 1.}
\label{table5}
\end{table}

\begin{table}[ht!]
\centering
\begin{tabular}{lrrr}
\toprule
            Game &      KFAC+Denoised & KFAC+TRL10x & Scattering\\
\midrule
           alien &       996 $\pm$ 180 &  643 $\pm$ 51 & 441 $\pm$ 90 \\
          amidar &          163 $\pm$ 15 &  98 $\pm$ 26 & 84 $\pm$ 11 \\
         assault &         501 $\pm$ 85& 496 $\pm$ 129 & 434 $\pm$ 304  \\
         asterix &          537 $\pm$ 96&  526 $\pm$ 64 & 502 $\pm$ 91 \\
      bank\_heist &              100 $\pm$ 14 & 57 $\pm$ 36 & 29 $\pm$ 13 \\
     battle\_zone &           8622 $\pm$ 5358  & 6156 $\pm$ 1951 & 4311 $\pm$ 1517 \\
          boxing &                0 $\pm$ 2& -1 $\pm$ 3 & -9 $\pm$  12  \\
        breakout &              3 $\pm$ 1& 2 $\pm$ 2 & 2 $\pm$  0  \\
 chopper\_command &            692 $\pm$ 81& 1302 $\pm$ 328 & 441 $\pm$ 80 \\
   crazy\_climber &         14242 $\pm$ 2936 & 3546 $\pm$ 1231 & 740 $\pm$ 291 \\
    demon\_attack &             582 $\pm$ 130 & 318 $\pm$ 168 & 692 $\pm$ 232 \\
         freeway &                26 $\pm$ 0& 24 $\pm$ 0 & 19 $\pm$ 1 \\
       frostbite &         1760 $\pm$ 448 & 1483 $\pm$ 466 & 654 $\pm$  709 \\
          gopher &         363 $\pm$ 4 & 265 $\pm$ 67 & 172 $\pm$ 3 \\
            hero &     4188 $\pm$ 1635 & 4206 $\pm$ 1862  & 4127 $\pm$ 1074 \\
       jamesbond &       263 $\pm$ 22 & 217 $\pm$ 68 & 48 $\pm$ 10 \\
        kangaroo &        2085 $\pm$ 2055 & 588 $\pm$ 5 & 391 $\pm$  52 \\
           krull &       2855 $\pm$ 156 & 3392 $\pm$ 2205 & 772$\pm$  560 \\
  kung\_fu\_master &      8481 $\pm$ 8270 & 7357 $\pm$ 9200 & 233$\pm$  205  \\
       ms\_pacman &        1137 $\pm$ 180 & 867 $\pm$ 128 & 613 $\pm$ 159   \\
            pong &        -19 $\pm$ 0.6&  -19 $\pm$ 1 & -20 $\pm$  0    \\
     private\_eye &      56 $\pm$ 42 &  100 $\pm$ 0 & 0 $\pm$ 0   \\
           qbert &        731 $\pm$ 256 & 538 $\pm$ 114 & 475 $\pm$ 161 \\
     road\_runner &      4516 $\pm$ 2869 & 7224 $\pm$ 4598 & 1278 $\pm$463  \\
        seaquest &       349 $\pm$ 63 & 520 $\pm$ 97 & 213 $\pm$ 96 \\
       up\_n\_down &      2557 $\pm$ 641 & 2108 $\pm$ 298 & 993 $\pm$ 244  \\
\bottomrule
\vspace{-12pt}
\end{tabular}
\label{table6}
\caption{Standard deviations across seeds for runs presented Table 2.}
\end{table}

\newpage
\subsection*{Further results and learning curves}

\textbf{Learning curves} As a simpler version of the experiments in the main text body, we show basic proof of concept on the simple \emph{Pong} Atari game. Our experimental setup consists in using our own implementation of \emph{prioritized double DQN} as a baseline, which combines algorithmic advances from \cite{prioritized} and \cite{doubledqn}. We replaced the densely connected layer of the original DQN architecture with a tensor regression layer implementing Tucker decomposition for different Tucker ranks, yielding different network compression factors. (These curves average three different random seeds).

\begin{figure}[ht]
\begin{center}
\includegraphics[width=0.99\linewidth]{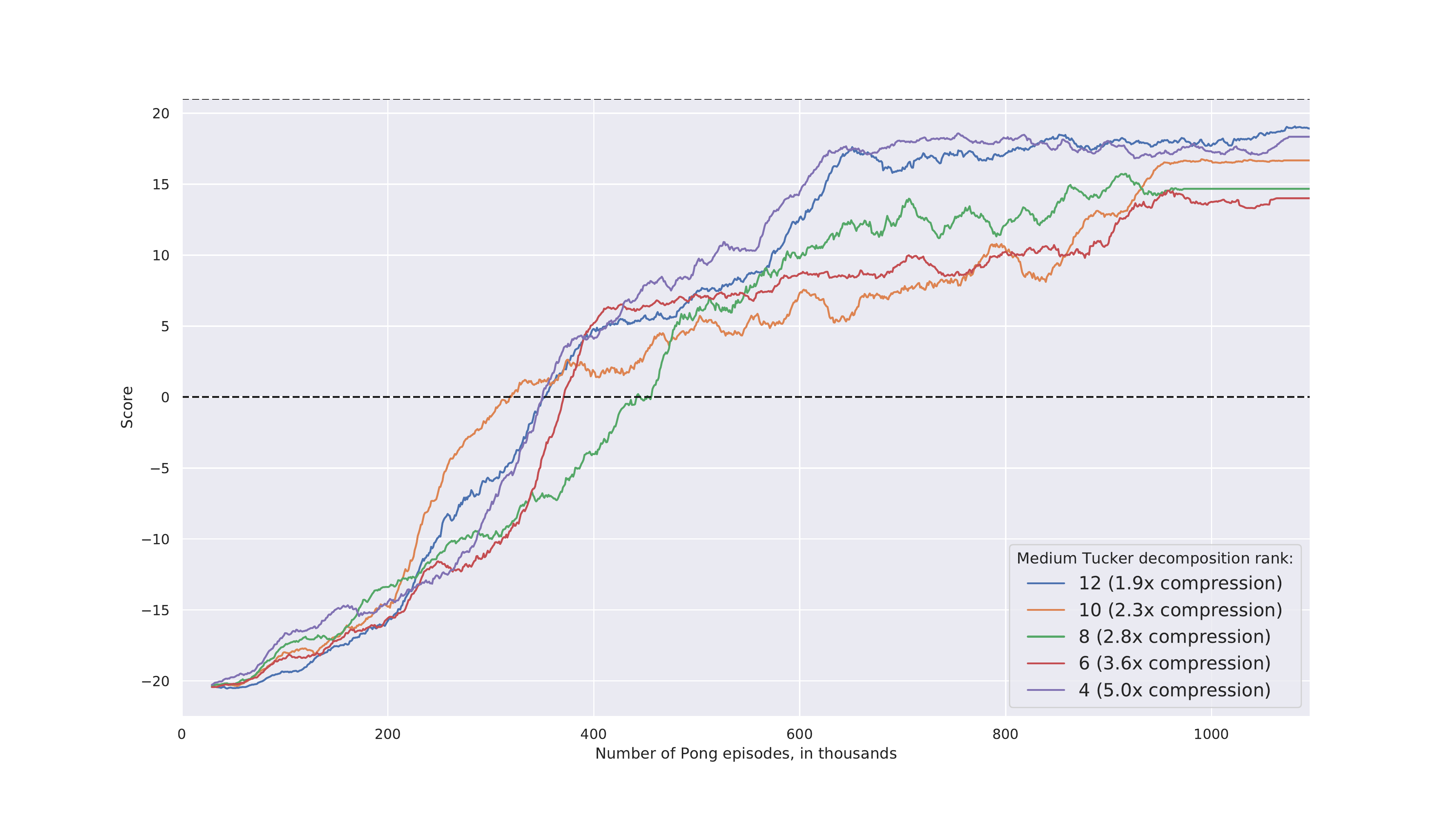}
\end{center}
\caption{Prioritized tensorized DQN on Atari Pong. Original learning curve versus several learning curves for five different Tucker ranks factorizations and therefore parameter compression rates (3 different random seeds each, with a 30 episodes moving average for legibility). Best viewed in colour.}
\label{fig:pong_prioritized_summary}
\end{figure}

\begin{figure}[ht]
\begin{center}
\includegraphics[width=0.7\linewidth]{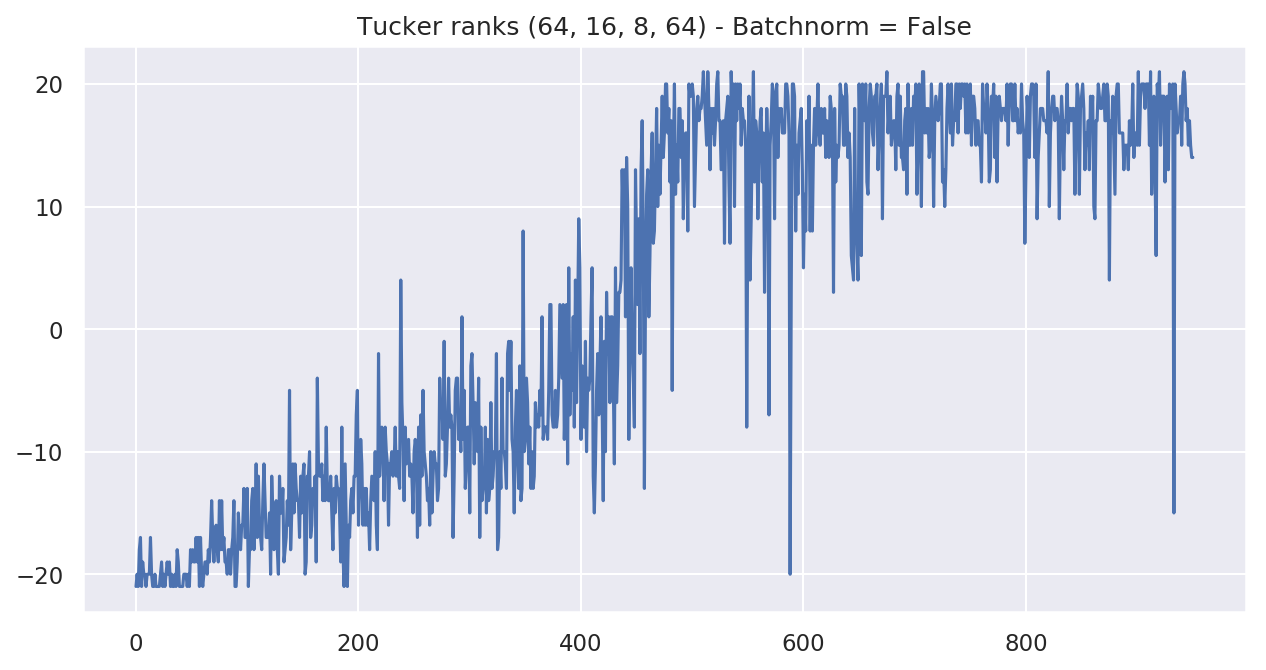}
\end{center}
\caption{Focus on a typical single run of the tensorized DQN learning. The overall shape of the typical learning curve is preserved, but drawdowns in the plateauing phase do appear.}
\label{fig:single_run_focus}
\end{figure}

\textbf{Qualitative behaviour.} First results, both in terms of learning performance and compression factor, can be seen in figure \ref{fig:pong_prioritized_summary}. The two main findings of this experiment are that first, and overall, the final performance of the agent remains unaffected by the tensor factorization, even with high compression rates nearing 10 times. Second, this obviously comes at the expense of stability during training - in tough compression regimes, learning curves are slightly delayed, and their plateauing phases contain occasional noisy drawdowns illustrating the increased difficulty of learning, as seen in figure \ref{fig:single_run_focus}. The extra pathwise noise, however, can be seen as promoting exploration.

\end{document}